\documentclass[10pt,twocolumn,letterpaper]{article}

\usepackage{wacv}
\usepackage{times}
\usepackage{epsfig}
\usepackage{graphicx}
\usepackage{amsmath}
\usepackage{amssymb}
\usepackage{booktabs}
\usepackage{multirow}
\usepackage{float}
\usepackage{subfig}

\usepackage{array}
\usepackage{xcolor, colortbl}

\def\wacvPaperID{1918} 

\wacvapplicationstrack 

\wacvfinalcopy 

\ifwacvfinal
\usepackage[breaklinks=true,bookmarks=false]{hyperref}
\else
\usepackage[pagebackref=true,breaklinks=true,colorlinks,bookmarks=false]{hyperref}
\fi

\begin{document}

\title{Audio-Visual Efficient Conformer for Robust Speech Recognition}

\author{Maxime Burchi, Radu Timofte\\
Computer Vision Lab, CAIDAS, IFI, University of Würzburg, Germany\\
{\tt\small \{maxime.burchi,radu.timofte\}@uni-wuerzburg.de}}

\maketitle
\thispagestyle{empty}

\newcommand{\reslrstwo}{2.3}
\newcommand{\reslrsthree}{1.8}
\definecolor{lightgray}{gray}{0.95}

\begin{abstract}
   End-to-end Automatic Speech Recognition (ASR) systems based on neural networks have seen large improvements in recent years. The availability of large scale hand-labeled datasets and sufficient computing resources made it possible to train powerful deep neural networks, reaching very low Word Error Rate (WER) on academic benchmarks. However, despite impressive performance on clean audio samples, a drop of performance is often observed on noisy speech. In this work, we propose to improve the noise robustness of the recently proposed Efficient Conformer Connectionist Temporal Classification (CTC)-based architecture by processing both audio and visual modalities. We improve previous lip reading methods using an Efficient Conformer back-end on top of a ResNet-18 visual front-end and by adding intermediate CTC losses between blocks. We condition intermediate block features on early predictions using Inter CTC residual modules to relax the conditional independence assumption of CTC-based models. We also replace the Efficient Conformer grouped attention by a more efficient and simpler attention mechanism that we call patch attention. We experiment with publicly available Lip Reading Sentences 2 (LRS2) and Lip Reading Sentences 3 (LRS3) datasets. Our experiments show that using audio and visual modalities allows to better recognize speech in the presence of environmental noise and significantly accelerate training, reaching lower WER with 4 times less training steps. Our Audio-Visual Efficient Conformer (AVEC) model achieves state-of-the-art performance, reaching WER of \reslrstwo\% and \reslrsthree\% on LRS2 and LRS3 test sets. Code and pretrained models are available at~\href{https://github.com/burchim/AVEC}{https://github.com/burchim/AVEC}.
\end{abstract}
\section{Introduction}
\begin{figure}[tb]
        \centering
        \includegraphics[width=0.75\linewidth]{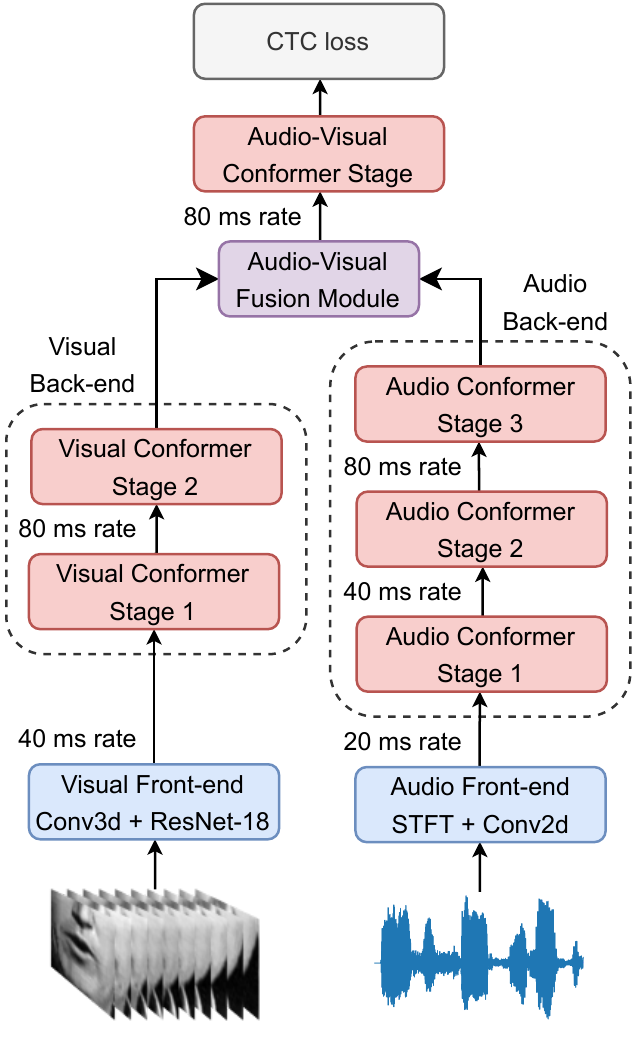}
        \caption{\textbf{Audio-Visual Efficient Conformer architecture.} The model is trained end-to-end using CTC loss and takes raw audio waveforms and lip movements from the speaker as inputs.}
    \label{fig:model}
    \vspace{-0.5cm}
\end{figure}

End-to-end Automatic Speech Recognition based on deep neural networks has become the standard of state-of-the-art approaches in recent years~\cite{kriman2020quartznet, zhang2020transformer, han2020contextnet, gulati2020conformer, guo2021recent, majumdar2021citrinet, burchi2021efficient}. The availability of large scale hand-labeled datasets and sufficient computing resources made it possible to train powerful deep neural networks for ASR, reaching very low WER on academic benchmarks like LibriSpeech~\cite{panayotov2015librispeech}. Neural architectures like Recurrent Neural Networks (RNN)~\cite{graves2013speech, hannun2014deep}, Convolution Neural Networks (CNN)~\cite{collobert2016wav2letter, li2019jasper} and Transformers~\cite{dong2018speech, karita2019comparative} have successfully been trained from raw audio waveforms and mel-spectrograms audio features to transcribe speech to text. Recently, Gulati~\textit{et al.}~\cite{gulati2020conformer} proposed a convolution-augmented transformer architecture (Conformer) to model both local and global dependencies using convolution and attention to reach better speech recognition performance. Concurrently, Nozaki~\textit{et al.}~\cite{nozaki2021relaxing} improved CTC-based speech recognition by conditioning intermediate encoder block features on early predictions using intermediate CTC losses~\cite{graves2006connectionist}. Burchi~\textit{et al.}~\cite{burchi2021efficient} also proposed an Efficient Conformer architecture using grouped attention for speech recognition, lowering the amount of computation while achieving better performance. Inspired from computer vision backbones, the Efficient Conformer encoder is composed of multiple stages where each stage comprises a number of Conformer blocks to progressively downsample and project the audio sequence to wider feature dimensions.

Yet, even if these audio-only approaches are breaking the state-of-the-art, one major pitfall for using them in the real-world is the rapid deterioration of performance in the presence of ambient noise. In parallel to that, Audio Visual Speech Recognition (AVSR) has recently attracted a lot of research attention due to its ability to use image processing techniques to aid speech recognition systems. Preceding works have shown that including the visual modality of lip movements could improve the robustness of ASR systems with respect to noise while reaching better recognition performance~\cite{son2017lip, sterpu2018attention, petridis2018audio, afouras2018deep, xu2020discriminative, ma2021end}. Xu~\textit{et al.}~\cite{xu2020discriminative} proposed a two-stage approach to first separate the target voice from background noise using the speakers lip movements and then transcribe the filtered audio signal with the help of lip movements. Petridis~\textit{et al.}~\cite{petridis2018audio} uses a hybrid architecture, training an LSTM-based sequence-to-sequence (S2S) model with an auxiliary CTC loss using an early fusion strategy to reach better performance. Ma~\textit{et al.}~\cite{ma2021end} uses Conformer back-end networks with ResNet-18~\cite{he2016deep} front-end networks to improve recognition performance. 
 
Other works focus on Visual Speech Recognition (VSR), only using lip movements to transcribe spoken language into text~\cite{assael2016lipnet, chung2017lip, zhang2019spatio, afouras2020asr, zhao2020hearing, prajwal2022sub, ma2022visual}. An important line of research is the use of cross-modal distillation. Afouras~\textit{et al.}~\cite{afouras2020asr} and Zhao~\textit{et al.}~\cite{zhao2020hearing} proposed to improve the lip reading performance by distilling from an ASR model trained on a large-scale audio-only corpus while Ma~\textit{et al.}~\cite{ma2022visual} uses prediction-based auxiliary tasks. Prajwal~\textit{et al.}~\cite{prajwal2022sub} also proposed to use sub-words units instead of characters to transcribe sequences, greatly reducing running time and memory requirements. Also providing a language prior, reducing the language modelling burden of the model.   

In this work we focus on the design of a noise robust speech recognition architecture processing both audio and visual modalities. We use the recently proposed CTC-based Efficient Conformer architecture~\cite{burchi2021efficient} and show that including the visual modality of lip movements can successfully improve noise robustness while significantly accelerating training. Our Audio-Visual Efficient Conformer (AVEC) reaches lower WER using 4 times less training steps than its audio-only counterpart. Moreover, we are the first work to apply intermediate CTC losses between blocks~\cite{lee2021intermediate, nozaki2021relaxing} to improve visual speech recognition performance. We show that conditioning intermediate features on early predictions using Inter CTC residual modules allows to close the gap in WER between autoregressive and non-autoregressive AVSR systems based on S2S. This also helps to counter a common failure case which is that audio-visual models tend to ignore the visual modality. In this way, we force pre-fusion layers to learn spatiotemporal features. Finally, we replace the Efficient Conformer grouped attention by a more efficient and simpler attention mechanism that we call patch attention. Patch attention reaches similar performance to grouped attention while having a lower complexity. The contributions of this work are as follows:
\begin{itemize}
    \item We improve the noise robustness of the recently proposed Efficient Conformer architecture by processing both audio and visual modalities.
    \item We condition intermediate Conformer block features on early predictions using Inter CTC residual modules to relax the conditional independence assumption of CTC models. This allows us to close the gap in WER between autoregressive and non-autoregressive methods based on S2S.
    \item We propose to replace the Efficient Conformer grouped attention by a more efficient and simpler attention mechanism that we call patch attention. Patch attention reaches similar performance to grouped attention with a lower complexity.
    \item We experiment on publicly available LRS2 and LRS3 datasets and reach state-of-the-art results using audio and visual modalities.
\end{itemize}

\section{Method}
In this section, we describe our proposed Audio-Visual Efficient Conformer network. The model is composed of 4 main components: An audio encoder, a visual encoder, an audio-visual fusion module and an audio-visual encoder. The audio and visual encoders are separated into modality specific front-end networks to transform each input modality into temporal sequences and Efficient Conformer back-end networks to model local and global temporal relationships. The full model is trained end-to-end using intermediate CTC losses between Conformer blocks in addition to the output CTC layer. The complete architecture of the model is shown in Figure~\ref{fig:model}. 

\subsection{Model Architecture}

\textbf{Audio front-end.} The audio front-end network first transforms raw audio wave-forms into mel-spectrograms using a short-time Fourier transform computed over windows of 20ms with a step size of 10ms. 80-dimensional mel-scale log filter banks are applied to the resulting frequency features. The mel-spectrograms are processed by a 2D convolution stem to extract local temporal-frequency features, resulting in a 20ms frame rate signal. The audio front-end architecture is shown in Table~\ref{table:audio-front}.

\begin{table}[ht]
\centering
\scriptsize
\caption{Audio Front-end architecture, 1.2 Millions parameters. $T_{a}$ denotes the input audio sample length.}
\begin{tabular}[t]{c|cc}
\hline
\multirow{2}{*}{Stage} & \multirow{2}{*}{Layers} & \multirow{2}{*}{Output Shape} \\\\ 
\hline\hline
$\begin{matrix}$Fourier$\\$Transf$\end{matrix}$& $\begin{matrix}$STFT: 400 window length$\\$160 hop length, 512 ffts$\end{matrix}$ & $(257,~T_{a} // 160 + 1)$\\
\hline
\multirow{2}{*}{\begin{tabular}[c]{@{}c@{}}Mel\\Scale\end{tabular}} & \multirow{2}{*}{Mel Scale: 80 mels} & \multirow{2}{*}{$(80,~T_{a} // 160 + 1)$} \\\\
\hline
\multirow{2}{*}{Stem} & \multirow{2}{*}{Conv2d: $3^{2}$, 180 filters, $2^{2}$ stride} & \multirow{2}{*}{$(180,~40,~T_{a} // 320 +1)$}  \\\\
\hline
\multirow{2}{*}{Proj}  & \multirow{2}{*}{Linear, 180 units} & \multirow{2}{*}{$(T_{a} // 320 +1,~180)$}\\\\
\hline
\end{tabular}
\label{table:audio-front}
\vspace{-0.25cm}
\end{table}

\textbf{Visual front-end.} The visual front-end network~\cite{ma2021end} transforms input video frames into temporal sequences. A 3D convolution stem with kernel size $5\times7\times7$ is first applied to the video. Each video frame is then processed independently using a 2D ResNet-18~\cite{he2016deep} with an output spatial average pooling. Temporal features are then projected to the back-end network input dimension using a linear layer. The visual front-end architecture is shown in Table~\ref{table:visual-front}.
\begin{table}[ht]
\centering
\scriptsize
\caption{Visual Front-end architecture, 11.3 Millions parameters. $T_{v}$ denotes the number of input video frames.}
\begin{tabular}[t]{c|cc}
\hline
\multirow{2}{*}{Stage} & \multirow{2}{*}{Layers} & \multirow{2}{*}{Output Shape}\\\\
\hline\hline
Stem & $\begin{matrix}$Conv3d: $5 \times 7^{2}$, 64 filters, $1 \times 2^{2}$ stride$\\$MaxPoo3d: $1 \times 3^{2}$, $1 \times 2^{2}$ stride$\end{matrix}$ & $(64,~T_{v},~22,~22)$\\
\hline
Res 1 & $2\times\begin{bmatrix}$Conv2d: $3^{2}$, 64 filters$\\$Conv2d: $3^{2}$, 64 filters$\end{bmatrix}$ & $(T_{v},~64,~22,~22)$\\
\hline
Res 2 & $2\times\begin{bmatrix}$Conv2d: $3^{2}$, 128 filters$\\$Conv2d: $3^{2}$, 128 filters$\end{bmatrix}$ & $(T_{v},~128,~11,~11)$\\
\hline
Res 3 & $2\times\begin{bmatrix}$Conv2d: $3^{2}$, 256 filters$\\$Conv2d: $3^{2}$, 256 filters$\end{bmatrix}$ & $(T_{v},~256,~6,~6)$\\
\hline
Res 4 & $2\times\begin{bmatrix}$Conv2d: $3^{2}$, 512 filters$\\$Conv2d: $3^{2}$, 512 filters$\end{bmatrix}$ & $(T_{v},~512,~3,~3)$\\
\hline
\multirow{2}{*}{Pool}  & \multirow{2}{*}{Global Average Pooling} & \multirow{2}{*}{$(T_{v},~512)$}\\\\
\hline
\multirow{2}{*}{Proj}  & \multirow{2}{*}{Linear, 256 units} & \multirow{2}{*}{$(T_{v},~256)$}\\\\
\hline
\end{tabular}
\label{table:visual-front}
\vspace{-0.25cm}
\end{table}

\textbf{Back-end networks.} The back-end networks use an Efficient Conformer architecture. The Efficient Conformer encoder was proposed in~\cite{burchi2021efficient}, it is composed of several stages where each stage comprises a number of Conformer blocks~\cite{gulati2020conformer} using grouped attention with relative positional encodings. The temporal sequence is progressively downsampled using strided convolutions and projected to wider feature dimensions, lowering the amount of computation while achieving better performance. We use 3 stages in the audio back-end network to downsample the audio signal to a 80 milliseconds frame rate. Only 2 stages are necessary to downsample the visual signal to the same frame rate. Table~\ref{table:back-end} shows the hyper-parameter of each back-end network.

\begin{table}[htb]
\centering
\caption{Back-end networks hyper-parameters. \textit{InterCTC blocks} indicates Conformer blocks having a post Inter CTC residual module.}
\scriptsize
\begin{tabular}{c|ccc}
\hline
\multirow{2}{*}{Network} & \multicolumn{1}{c}{\multirow{2}{*}{\begin{tabular}[c]{@{}c@{}}Visual\\ Back-end\end{tabular}}} & \multicolumn{1}{c}{\multirow{2}{*}{\begin{tabular}[c]{@{}c@{}}Audio\\ Back-end\end{tabular}}} & \multicolumn{1}{c}{\multirow{2}{*}{\begin{tabular}[c]{@{}c@{}}Audio-Visual\\ Encoder\end{tabular}}} \\ \\ \hline\hline
Num Params (M) & 13.6          & 17.9            & 15.9              \\
Num Stages & 2 & 3 & 1\\
Blocks per Stage   & 6, 1            & 5, 6, 1           & 5             \\
Total Num Blocks   & 7            & 12           & 5             \\
Stage Feature Dim  & 256, 360           & 180, 256, 360     & 360         \\
Conv Kernel Size & 15 & 15 & 15 \\
Stage Patch Size  & 1, 1             & 3, 1, 1          & 1             \\
InterCTC Blocks  & 3, 6             & 8, 11          & 2             \\\hline
\end{tabular}
\label{table:hp}
\end{table}

\textbf{Audio-visual fusion module.} Similar to~\cite{petridis2018audio, ma2021end}, we use an early fusion strategy to learn audio-visual features and reduce model complexity. The acoustic and visual features from the back-end networks are concatenated and fed into a joint feed-forward network. The concatenated features of size $2 \times d_{model}$ are first expanded using a linear layer with output size $d_{ff} = 4 \times d_{model}$, passed through a Swish activation function~\cite{ramachandran2017searching} and projected back to the original feature dimension $d_{model}$.

\textbf{Audio-visual encoder.} The audio-visual encoder is a single stage back-end network composed of 5 Conformer blocks without downsampling. The encoder outputs are then projected to a CTC layer to maximize the sum of probabilities of correct target alignments.

\begin{figure*}[tb]
    \begin{minipage}[]{1.0\linewidth}
        \centering
        \centerline{\includegraphics[width=\linewidth]{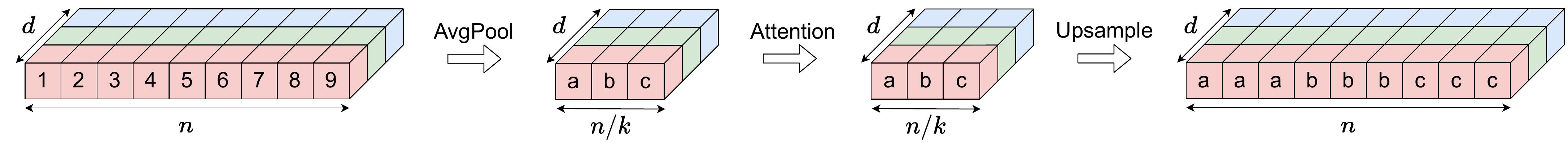}}
        \caption{\textbf{Patch Multi-Head Self-Attention.} The input sequence is downsampled using an average pooling before applying multi-head self-attention. The output sequence is then upsampled via nearest neighbor upsampling, reducing attention complexity from $O(n^{2}\cdot d)$ to $O((n/k)^{2}\cdot d)$ where $k$ defines the pooling / upsampling kernel size. Patch attention is equivalent to regular attention when $k=1$.}
        \label{fig:PatchAtt}
    \end{minipage}
\end{figure*}

\subsection{Patch Attention.}
The Efficient Conformer~\cite{burchi2021efficient} proposed to replace Multi-Head Self-Attention (MHSA)~\cite{vaswani2017attention} in earlier encoder layers with grouped attention. Grouped MHSA reduce attention complexity by grouping neighbouring temporal elements along the feature dimension before applying scaled dot-product attention. Attention having a quadratic computational complexity with respect to the sequence length, this caused the network to have an asymmetric complexity with earlier attention layers requiring more flops than latter layers with shorter sequence length. In this work, we propose to replace grouped attention with a simpler and more efficient attention mechanism that we call patch attention (Figure~\ref{fig:PatchAtt}). 
\begin{table}[b]
\centering
\scriptsize
\caption{Attention variants complexities including query, key, value and output linear projections. $n$ and $d$ are the sequence length and feature dimension respectively.}
\hfill \break
\begin{tabular}{c|cc}
\hline
\multirow{2}{*}{\begin{tabular}[c]{@{}c@{}}Attention \\ Variant\end{tabular}} & \multirow{2}{*}{\begin{tabular}[c]{@{}c@{}}Hyper \\ Parameter\end{tabular}} & \multirow{2}{*}{\begin{tabular}[c]{@{}c@{}}Full Attention \\ Complexity\end{tabular}} \\\\ \hline\hline
Regular & - & $O(n \cdot d^{2}+n^{2} \cdot d)$          \\
Grouped & Group Size (g) & $O(n \cdot d^{2}+(n/g)^{2}\cdot d \cdot g)$            \\
Patch & Patch Size (k) & $O(n / k \cdot d^{2}+(n/k)^{2} \cdot d)$            \\\hline
\end{tabular}
\label{table:complexity}
\end{table}
Similar to the pooling attention proposed by the Multiscale Vision Transformer (MViT)~\cite{fan2021multiscale} for video and image recognition, the patch attention proceed to an average pooling on the input sequence before projection the query, key and values. 
\begin{align}
& X = AvgPooling1d(X_{in})\\
&\text{with}\ Q, K, V=XW^{Q}, XW^{K}, XW^{V}
\end{align}
Where $W^{Q}$, $W^{K}$, $W^{V}\in \mathbb{R}^{d \times d}$ are query, key and value linear projections parameter matrices. MHSA with relative sinusoidal positional encoding is then performed at lower resolution as: 
\begin{align}
& MHSA(X) = Concat\left(O_{1}, ..., O_{H}\right)W^{O}\\
&\text{with}\ O_{h} = softmax\left(\frac{Q_{h}K_{h}^{T}+S^{rel}_{h}}{\sqrt{d_{h}}}\right)V_{h}
\end{align}
Where $S^{rel} \in \mathbb{R}^{n \times n}$ is a relative position score matrix that satisfy $S^{rel}[i, j]=Q_{i}E_{j-i}^{T}$. $E$ is the linear projection of a standard sinusoidal positional encoding matrix with positions ranging from $-(n_{max}-1)$ to $(n_{max}-1)$. The attention output sequence is then projected and up-sampled back to the initial resolution using nearest neighbor up-sampling.
\begin{align}
X_{out} = UpsampleNearest1d(MHSA(X))
\end{align}
\begin{figure}[b]
        \centering
        \includegraphics[width=1.0\linewidth]{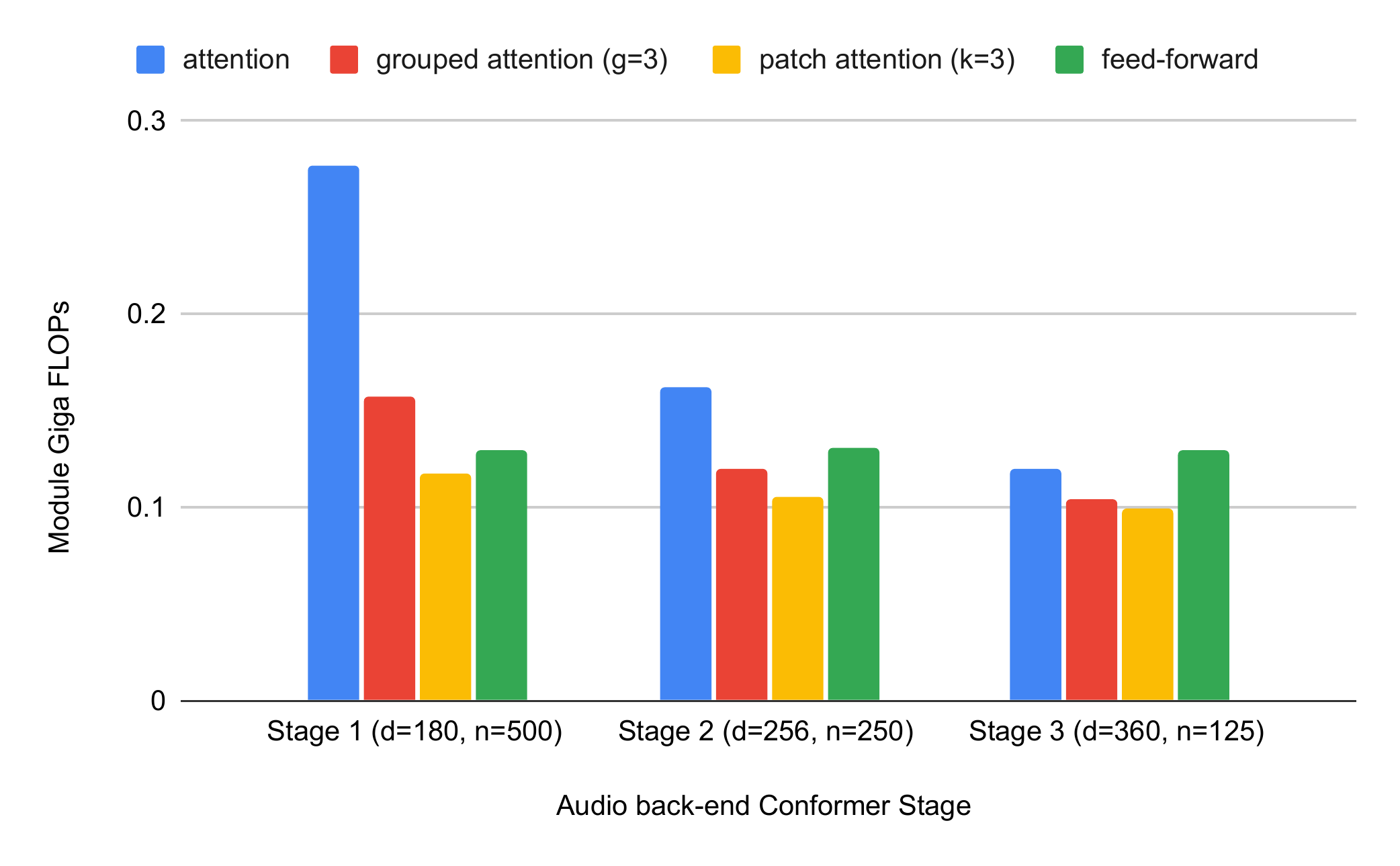}
        \caption{\textbf{Audio-only back-end modules FLOPs (Billion).}}
    \label{fig:module_flops}
    \vspace{-0.5cm}
\end{figure}
In consequence, each temporal element of the same patch produce the same attention output. Local temporal relationships are only modeled in the convolution modules while global relationships are modeled by patch attention. We use 1-dimensional patches in this work but patch attention could also be generalized to image and video data using 2D and 3D patches. We leave this to future works. The computational complexity of each attention variant is shown in Table~\ref{table:complexity}. Path attention further reduce complexity compared to grouped attention by decreasing the amount of computation needed by Query, Key, Value and Output fully connected layers while keeping the feature dimension unchanged. Similar to previous work~\cite{burchi2021efficient}, we only use patch attention in the first audio back-end stage to reduce complexity while maintaining model recognition performance. Figure~\ref{fig:module_flops} shows the amount of FLOPs for each attention module variant with respect to encoded sequence length $n$ and model feature dimension $d$. Using patch or grouped attention variants instead of regular MHSA greatly reduce the amount of FLOPs in the first audio back-end stage.

\subsection{Intermediate CTC Predictions.} 
\begin{figure}[tb]
        \centering
        \includegraphics[width=0.75\linewidth]{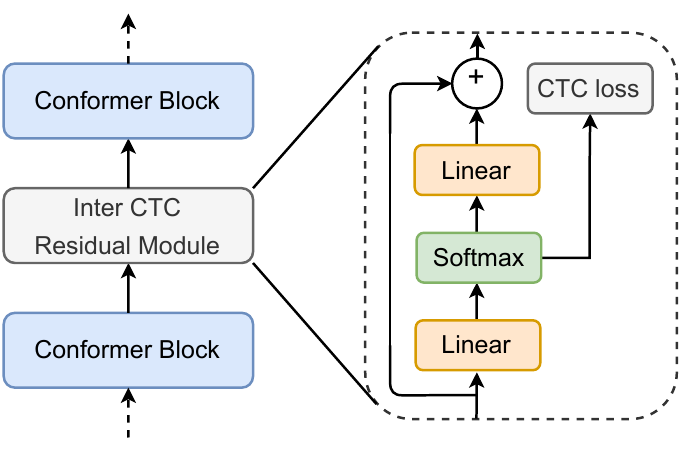}
        \caption{\textbf{Inter CTC residual module.} Intermediate predictions are summed to the input of the next Conformer block to condition the prediction of the final block on it. Intermediate CTC losses are added to the output CTC loss for the computation of the final loss.}
    \label{fig:interCTC}
    \vspace{-0.25cm}
\end{figure}

Inspired by \cite{lee2021intermediate} and \cite{nozaki2021relaxing} who proposed to add intermediate CTC losses between encoder blocks to improve CTC-based speech recognition performance, we add Inter CTC residual modules (Figure~\ref{fig:interCTC}) in encoder networks. We condition intermediate block features of both audio, visual and audio-visual encoders on early predictions to relax the conditional independence assumption of CTC models. During both training and inference, each intermediate prediction is summed to the input of the next layer to help recognition. We use the same method proposed in~\cite{nozaki2021relaxing} except that we do not share layer parameters between losses. The $l^{th}$ block output $X^{out}_{l}$ is passed through a feed-forward network with residual connection and a softmax activation function:
\begin{align}
& Z_{l}=Softmax(Linear(X^{out}_{l}))\\
& X^{in}_{l+1}=X^{out}_{l}+Linear(Z_{l})
\end{align}
Where $Z_{l} \in \mathbb{R}^{T \times V}$ is a probability distribution over the output vocabulary. The intermediate CTC loss is then computed using the target sequence $y$ as:
\begin{align}
& L^{inter}_{l}=-log(P(y|Z_{l}))\\
&\text{with}\ P(y|Z_{l}) = \sum_{\pi \in \mathcal{B}_{CTC}^{-1}(y)}\prod_{t=1}^{T}Z_{t,\pi_{t}}
\end{align}
Where $\pi \in V^{T}$ are paths of tokens and $\mathcal{B}_{CTC}$ is a many-to-one map that simply removes all blanks and repeated labels from the paths. The total training objective is defined as follows:
\begin{flalign}
& L = (1 - \lambda)L^{CTC} + \lambda L^{inter}\\
&\text{with}\ L^{inter} = \frac{1}{K}\sum_{k \in \textit{interblocks}}{L^{inter}_{k}}
\end{flalign}
Where \textit{interblocks} is the set of blocks having a post Inter CTC residual module (Figure~\ref{fig:interCTC}). Similar to \cite{nozaki2021relaxing}, we use Inter CTC residual modules every 3 Conformer blocks with $\lambda$ set to 0.5 in every experiments.
  
\section{Experiments}

\subsection{Datasets}

We use 3 publicly available AVSR datasets in this work. The Lip Reading in the Wild (LRW)~\cite{chung2016lip} dataset is used for visual pre-training and the Lip Reading Sentences 2 (LRS2)~\cite{afouras2018deep} and Lip Reading Sentences 3 (LRS3)~\cite{afouras2018lrs3} datasets are used for training and evaluation.

\textbf{LRW dataset.} LRW is an audio-visual word recognition dataset consisting of short video segments containing a single word out of a vocabulary of 500. The dataset comprise 488,766 training samples with at least 800 utterances per class and a validation and test sets of 25,000 samples containing 50 utterances per class.

\textbf{LRS2 \& LRS3 datasets.} The LRS2 dataset is composed of 224.1 hours with 144,482 videos clips from the BBC television whereas the LRS3 dataset consists of 438.9 hours with 151,819 video clips extracted from TED and TEDx talks. Both datasets include corresponding subtitles with word alignment boundaries and are composed of a pre-train split, train-val split and test split. LRS2 has 96,318 utterances for pre-training (195 hours), 45,839 for training (28 hours), 1,082 for validation (0.6 hours), and 1,243 for testing (0.5 hours). Whereas LRS3 has 118,516 utterances in the pre-training set (408 hours), 31,982 utterances in the training-validation set (30 hours) and 1,321 utterances in the test set (0.9 hours). All videos contain a single speaker, have a $224\times 224$ pixels resolution and are sampled at 25 fps with 16kHz audio.

\begin{table*}[ht]
\centering
\caption{Comparison of WER (\%) on LRS2 / LRS3 test sets with recently published methods using publicly and non-publicly available datasets for Audio-Only (AO), Visual-Only (VO) and Audio-Visual (AV) models.}
\hfill \break
\begin{tabular}{ccccccc}
\hline
\multirow{2}{*}{\textbf{Method}} & \multicolumn{1}{c}{\multirow{2}{*}{\begin{tabular}[c]{@{}c@{}}\textbf{Model} \\ \textbf{Criterion}\end{tabular}}} & \multicolumn{1}{c}{\multirow{2}{*}{\begin{tabular}[c]{@{}c@{}}\textbf{Training} \\ \textbf{Datasets}\end{tabular}}} & \multicolumn{1}{c}{\multirow{2}{*}{\begin{tabular}[c]{@{}c@{}}\textbf{Total} \\ \textbf{Hours}\end{tabular}}} & \multicolumn{3}{c}{\textbf{test WER}}\\ 
& & & & \textbf{AO} & \textbf{VO} & \textbf{AV} \\\hline\hline

\multicolumn{7}{c}{($\downarrow$) \textit{Using Publicly Available Datasets} ($\downarrow$)}\\ \hline

Petridis~\textit{et al.}~\cite{petridis2018audio} & CTC+S2S & LRW, LRS2 & 381 & 8.3 / - & 63.5 / - & 7.0 / -\\ \hline

Zhang~\textit{et al.}~\cite{zhang2019spatio} &  S2S & LRW, LRS2\&3 & 788 / 790 & - & 51.7 / 60.1 & - \\ \hline

Afouras~\textit{et al.}~\cite{afouras2020asr} & CTC & VoxCeleb2\textsuperscript{clean}, LRS2\&3 & 1,032 / 808 & - & 51.3 / 59.8 & -\\\hline

Xu~\textit{et al.}~\cite{xu2020discriminative} & S2S & LRW, LRS3 & 595 & - / 7.2 & - / 57.8 & - / 6.8 \\\hline

Yu~\textit{et al.}\cite{yu2020audio} & LF-MMI & LRS2 & 224 & 6.7 / - & 48.9 / - & 5.9 / - \\ \hline

Ma~\textit{et al.}~\cite{ma2021end} & CTC+S2S & LRW, LRS2\&3 & 381 / 595 & 3.9 / 2.3 & 37.9 / 43.3 & 3.7 / 2.3 \\\hline

Prajwal~\textit{et al.}~\cite{prajwal2022sub} & S2S & LRS2\&3 & 698 & - & 28.9 / 40.6 & - \\\hline

Ma~\textit{et al.}~\cite{ma2022visual} & CTC+S2S & LRW, LRS2\&3  & 818 & - & \textbf{27.3 / 34.7} & - \\\hline

\rowcolor{lightgray}\textbf{Ours} & CTC & LRW, LRS2\&3 & 818 & \textbf{2.8 / 2.1} & 32.6 / 39.2 & \textbf{2.5 / 1.9} \\\hline

\rowcolor{lightgray}\textbf{+ Neural LM} & CTC & LRW, LRS2\&3 & 818 & \textbf{2.4 / 2.0} & 29.8 / 37.5 & \textbf{\reslrstwo~/~\reslrsthree} \\\hline\hline

\multicolumn{7}{c}{($\downarrow$) \textit{Using Non-Publicly Available Datasets} ($\downarrow$)}\\ \hline

Afouras~\textit{et al.}~\cite{afouras2018deep} & S2S & MVLRS, LRS2\&3 & 1,395 & 9.7 / 8.3 & 48.3 / 58.9 & 8.5 / 7.2\\\hline

Zhao~\textit{et al.}~\cite{zhao2020hearing} & S2S & MVLRS, LRS2 & 954 & - & 65.3 / - & - \\ \hline

Shillingford~\textit{et al.}~\cite{shillingford2018large} & CTC & LRVSR & 3,886 & - & - / 55.1 & -\\ \hline

Makino~\textit{et al.}~\cite{makino2019recurrent} & Transducer & YouTube-31k & 31,000 & - / 4.8 & - / 33.6 & - / 4.5 \\\hline

Serdyuk~\textit{et al.}~\cite{serdyuk2021audio} & Transducer & YouTube-90k & 91,000 & - & - / 25.9 & - / 2.3 \\\hline

Prajwal~\textit{et al.}~\cite{prajwal2022sub} & S2S & MVLRS, TEDx\textsubscript{ext}, LRS2\&3 & 2,676 & - & 22.6 / 30.7 & - \\\hline

Ma~\textit{et al.}~\cite{ma2022visual} & CTC+S2S & LRW, AVSpeech, LRS2\&3  & 1,459 & - & 25.5 / 31.5 & - \\\hline

\end{tabular}
\label{table:results}
\vspace{-0.25cm}
\end{table*}

\subsection{Implementation Details}
\textbf{Pre-processing} Similar to~\cite{ma2021end}, we remove differences related to rotation and scale by cropping the lip regions using bounding boxes of $96 \times 96$ pixels to facilitate recognition. The RetinaFace~\cite{deng2020retinaface} face detector and Face Alignment Network (FAN)~\cite{bulat2017far} are used to detect 68 facial landmarks. The cropped images are then converted to gray-scale and normalised between $-1$ and $1$. Facial landmarks of the LRW, LRS2 and LRS3 datasets are obtained from previous work~\cite{ma2022visual} and reused for pre-processing to get a clean comparison of the methods. A byte-pair encoding tokenizer is built from LRS2\&3 pre-train and trainval splits transcripts using sentencepiece~\cite{kudo2018sentencepiece}. We use a vocabulary size of 256 including the CTC blank token following preceding works on CTC-based speech recognition~\cite{majumdar2021citrinet,burchi2021efficient}.

\textbf{Data augmentation} Spec-Augment~\cite{park2020specaugment} is applied on the audio mel-spectrograms during training to prevent over-fitting with two frequency masks with mask size parameter $F=27$ and five time masks with adaptive size $pS=0.05$. Similarly to~\cite{ma2022visual}, we mask videos on the time axis using one mask per second with the maximum mask duration set to 0.4 seconds. Random cropping with size $88 \times 88$ and horizontal flipping are also performed for each video during training. We also follow Prajwal~\textit{et al.}~\cite{prajwal2022sub} using central crop with horizontal flipping at test time for visual-only experiments.

\textbf{Training Setup} We first pre-train the visual encoder on the LRW dataset~\cite{chung2016lip} using cross-entropy loss to recognize words being spoken. The visual encoder is pre-trained for 30 epochs and front-end weights are then used as initialization for training. Audio and visual encoders are trained on the LRS2\&3 datasets using a Noam schedule~\cite{vaswani2017attention} with 10k warmup steps and a peak learning rate of 1e-3. We use the Adam optimizer~\cite{kingma2014adam} with $\beta_1=0.9$, $\beta_2=0.98$. L2 regularization with a 1e-6 weight is also added to all the trainable weights of the model. We train all models with a global batch size of 256 on 4 GPUs, using a batch size of 16 per GPU with 4 accumulated steps. Nvidia A100 40GB GPUs are used for visual-only and audio-visual experiments while RTX 2080 Ti are used for audio-only experiments. The audio-only models are trained for 200 epochs while visual-only and audio-visual models are trained for 100 and 70 epochs respectively. Note that we only keep videos shorter than 400 frames (16 seconds) during training. Finally, we average models weights over the last 10 epoch checkpoints using Stochastic Weight Averaging~\cite{izmailov2018averaging} before evaluation. 

\textbf{Language Models.} Similarly to~\cite{li2019jasper}, we experiment with a N-gram~\cite{heafield2011kenlm} statistical language model (LM) and a Transformer neural language model. A 6-gram LM is used to generate a list of hypotheses using beam search and an external Transformer LM is used to rescore the final list. The 6-gram LM is trained on the LRS2\&3 pre-train and train-val transcriptions. Concerning the neural LM, we pretrain a 12 layer GPT-3 Small~\cite{brown2020language} on the LibriSpeech LM corpus for 0.5M steps using a batch size of 0.1M tokens and finetune it for 10 epochs on the LRS2\&3 transcriptions.

\subsection{Results}
Table~\ref{table:results} compares WERs of our Audio-Visual Efficient Conformer with state-of-the-art methods on the LRS2 and LRS3 test sets. Our Audio-Visual Efficient Conformer achieves state-of-the-art performances with WER of \reslrstwo\%/\reslrsthree\%. On the visual-only track, our CTC model competes with most recent autoregressive methods using S2S criterion. We were able to recover similar results but still lack behind~\textit{Ma et al.}~\cite{ma2022visual} which uses auxiliary losses with pre-trained audio-only and visual-only networks. We found our audio-visual network to converge faster than audio-only experiments, reaching better performance using 4 times less training steps. The intermediate CTC losses of the visual encoder could reach lower levels than in visual-only experiments showing that optimizing audio-visual layers can help pre-fusion layers to learn better representations.

\subsection{Ablation Studies}
We propose a detailed ablation study to better understand the improvements in terms of complexity and WER brought by each architectural modification. We report the number of operations measured in FLOPs (number of multiply-and-add operations) for the network to process a ten second audio/video clip. Inverse Real Time Factor (Inv RTF) is also measured on the LRS3 test set by decoding with a batch size 1 on a single Intel Core i7-12700 CPU thread. All ablations were performed by training audio-only models for 200 epochs and visual-only / audio-visual models for 50 epochs.

\begin{figure*}[ht]
    \centering
    \includegraphics[width=0.95\linewidth]{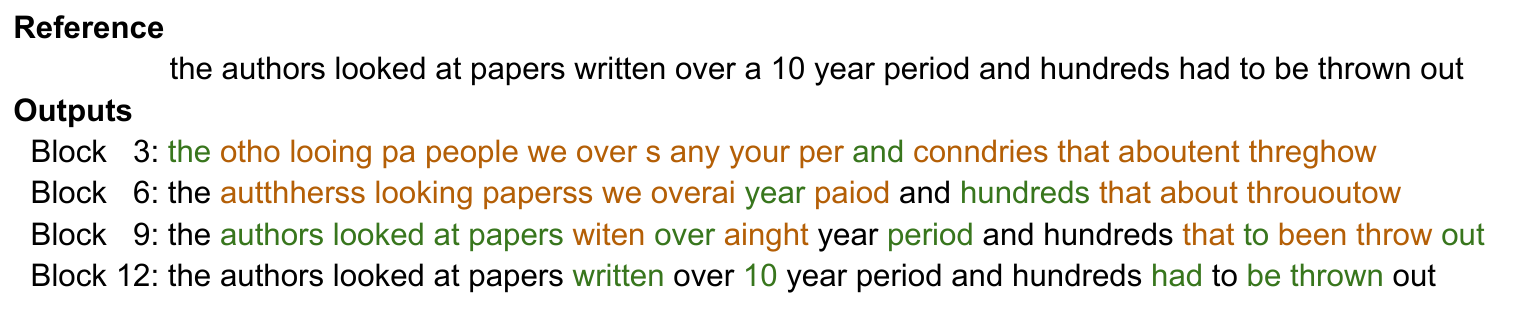}
    \caption{Output example of our Visual-only model using greedy search decoding on the LRS3 test set with intermediate CTC prediction every 3 blocks. The sentence is almost correctly transcribed except for the missing 'a' before '10 year'.}
    \label{fig:inter_preds}
\end{figure*}

\textbf{Efficient Conformer Visual Back-end.} We improve the recently proposed visual Conformer encoder~\cite{ma2021end} using an Efficient Conformer back-end network. The use of byte pair encodings for tokenization instead of characters allows us to further downsample temporal sequences without impacting the computation of CTC loss. Table~\ref{table:back-end} shows that using an Efficient Conformer back-end network for our visual-only model leads to better performances while reducing model complexity and training time. The number of model parameters is also slightly decreased.
\begin{table}[ht]
\vspace{-0.25cm}
\centering
\scriptsize
\caption{Ablation study on visual back-end network.}
\hfill \break
\begin{tabular}{c|ccccc}
\hline
\multirow{2}{*}{\begin{tabular}[c]{@{}c@{}}\textbf{Visual} \\ \textbf{Back-end}\end{tabular}} & \multicolumn{1}{c}{\multirow{2}{*}{\begin{tabular}[c]{@{}c@{}}\textbf{\#Params} \\ \textbf{(Million)}\end{tabular}}} & \multirow{2}{*}{\begin{tabular}[c]{@{}c@{}}\textbf{LRS2}\\ \textbf{test}\end{tabular}} & \multirow{2}{*}{\begin{tabular}[c]{@{}c@{}}\textbf{LRS3}\\ \textbf{test}\end{tabular}} & \multirow{2}{*}{\begin{tabular}[c]{@{}c@{}}\textbf{\#FLOPs} \\ \textbf{(Billion)}\end{tabular}} & \multirow{2}{*}{\begin{tabular}[c]{@{}c@{}}\textbf{Inv} \\ \textbf{RTF}\end{tabular}}\\\\ \hline\hline
Conformer           & 43.0 & 39.53  & 47.14 & 87.94 & 5.17 \\
Eff Conf             & 40.4 & \textbf{37.39} & \textbf{44.96} & 84.52 & 5.26                           \\\hline
\end{tabular}
\label{table:back-end}
\end{table}

\textbf{Inter CTC residual modules.}
Similar to~\cite{nozaki2021relaxing}, we experiment adding Inter CTC residual modules between blocks to relax the conditional independence assumption of CTC. Table~\ref{table:interctc} shows that using intermediate CTC losses every 3 Conformer blocks greatly helps to reduce WER, except for the audio-only setting where this does not improve performance. Figure~\ref{fig:inter_preds} gives an example of intermediate block predictions decoded using greedy search without an external language model on the test set of LRS3. We can see that the output is being refined in the encoder layers by conditioning on the intermediate predictions of previous layers. Since our model refines the output over the frame-level predictions, it can correct insertion and deletion errors in addition to substitution errors. We further study the impact of Inter CTC on multi-modal learning by measuring the performance of our audio-visual model when one of the two modalities is masked. As pointed out by preceding works~\cite{chung2016lip, afouras2018deep, makino2019recurrent}, networks with multi-modal inputs can often be dominated by one of the modes. In our case speech recognition is a significantly easier problem than lip reading which can cause the model to ignore visual information. Table~\ref{table:interctc_mask} shows that Inter CTC can help to counter this problem by forcing pre-fusion layers to transcribe the input signal.
\begin{table}[ht]
\vspace{-0.25cm}
\centering
\scriptsize
\caption{Ablation study on Inter CTC residual modules.}
\hfill \break
\begin{tabular}{cccccc}
\hline
\multirow{2}{*}{\begin{tabular}[c]{@{}c@{}}\textbf{Model} \\ \textbf{Back-end}\end{tabular}} & \multicolumn{1}{c}{\multirow{2}{*}{\begin{tabular}[c]{@{}c@{}}\textbf{\#Params} \\ \textbf{(Million)}\end{tabular}}} & \multirow{2}{*}{\begin{tabular}[c]{@{}c@{}}\textbf{LRS2}\\ \textbf{test}\end{tabular}} & \multirow{2}{*}{\begin{tabular}[c]{@{}c@{}}\textbf{LRS3}\\ \textbf{test}\end{tabular}} & \multirow{2}{*}{\begin{tabular}[c]{@{}c@{}}\textbf{\#FLOPs} \\ \textbf{(Billion)}\end{tabular}} & \multirow{2}{*}{\begin{tabular}[c]{@{}c@{}}\textbf{Inv} \\ \textbf{RTF}\end{tabular}}\\\\ \hline\hline
\textit{Audio-only} ($\downarrow$) \\\hline
Eff Conf     & 31.5 & 2.83 & 2.13 & 7.54 & 51.98              \\\hline
+ Inter CTC    & 32.1 & 2.84 & 2.11 & 7.67 & 50.30 \\\hline\hline
\textit{Visual-only} ($\downarrow$) \\\hline
Eff Conf             & 40.4 & 37.39 & 44.96 & 84.52 & 5.26              \\\hline
+ Inter CTC    & 40.9 & \textbf{33.82} & \textbf{40.63} & 84.60 & 5.26 \\\hline\hline
\textit{Audio-visual} ($\downarrow$) \\\hline
Eff Conf             & 60.9 & 2.87 & 2.54 & 90.53 & 4.84             \\\hline
+ Inter CTC    & 61.7 & 2.58 & 1.99 & 90.66 & 4.82 \\\hline
\end{tabular}
\label{table:interctc}
\end{table}
\begin{table}[ht]
\vspace{-0.75cm}
\centering
\scriptsize
\caption{Impact of Inter CTC on audio-visual model WER (\%) for LRS2 / LRS3 test sets in a masked modality setting.}
\hfill \break
\begin{tabular}{c|ccc}
\hline
\multirow{2}{*}{\textbf{Inter CTC}}  & \multicolumn{3}{c}{\textbf{Audio-Visual Eval Mode}}\\
& \textbf{masked video} & \textbf{masked audio} & \textbf{no mask}\\ \hline\hline
No & 4.48 / 3.22 & 52.77 / 59.10 & 2.87 / 2.54 \\
Yes & \textbf{3.39 / 2.38} & \textbf{37.62 / 46.55} & \textbf{2.58 / 1.99} \\
\hline
\end{tabular}
\label{table:interctc_mask}
\end{table}

\textbf{Patch multi-head self-attention.} We experiment replacing grouped attention by patch attention in the first audio encoder stage.  Our objective being to increase the model efficiency and simplicity without harming performance. Grouped attention was proposed in~\cite{burchi2021efficient} to reduce attention complexity for long sequences in the first encoder stage. Table~\ref{table:attention} shows the impact of each attention variant on our audio-only model performance and complexity. We start with an Efficient Conformer (M)~\cite{burchi2021efficient} and replace the attention mechanism. We find that grouped attention can be replaced by patch attention without a loss of performance using a patch size of 3 in the first back-end stage.
\begin{table}[ht]
\centering
\scriptsize
\caption{Ablation study on audio back-end attention.}
\hfill \break
\begin{tabular}{c|ccccc}
\hline
\multirow{2}{*}{\begin{tabular}[c]{@{}c@{}}\textbf{Attention} \\ \textbf{Type}\end{tabular}} & \multicolumn{1}{c}{\multirow{2}{*}{\begin{tabular}[c]{@{}c@{}}\textbf{Group /} \\ \textbf{Patch Size}\end{tabular}}} & \multirow{2}{*}{\begin{tabular}[c]{@{}c@{}}\textbf{LRS2}\\ \textbf{test}\end{tabular}} & \multirow{2}{*}{\begin{tabular}[c]{@{}c@{}}\textbf{LRS3}\\ \textbf{test}\end{tabular}} & \multirow{2}{*}{\begin{tabular}[c]{@{}c@{}}\textbf{\#FLOPs} \\ \textbf{(Billion)}\end{tabular}} & \multirow{2}{*}{\begin{tabular}[c]{@{}c@{}}\textbf{Inv} \\ \textbf{RTF}\end{tabular}}\\\\ \hline\hline
Regular & - & 2.85 & 2.12 & 8.66 & 49.86           \\
Grouped & 3, 1, 1 & 2.82 & 2.13 & 8.06 & 50.27            \\
Patch & 3, 1, 1   & 2.83 & 2.13 & \textbf{7.54}  & \textbf{51.98}               \\\hline
\end{tabular}
\label{table:attention}
\vspace{-0.25cm}
\end{table}

\subsection{Noise Robustness} 
\begin{figure}[ht]
    \begin{minipage}[]{1.0\linewidth}
        \subfloat[Babble noise]{\includegraphics[width=\textwidth]{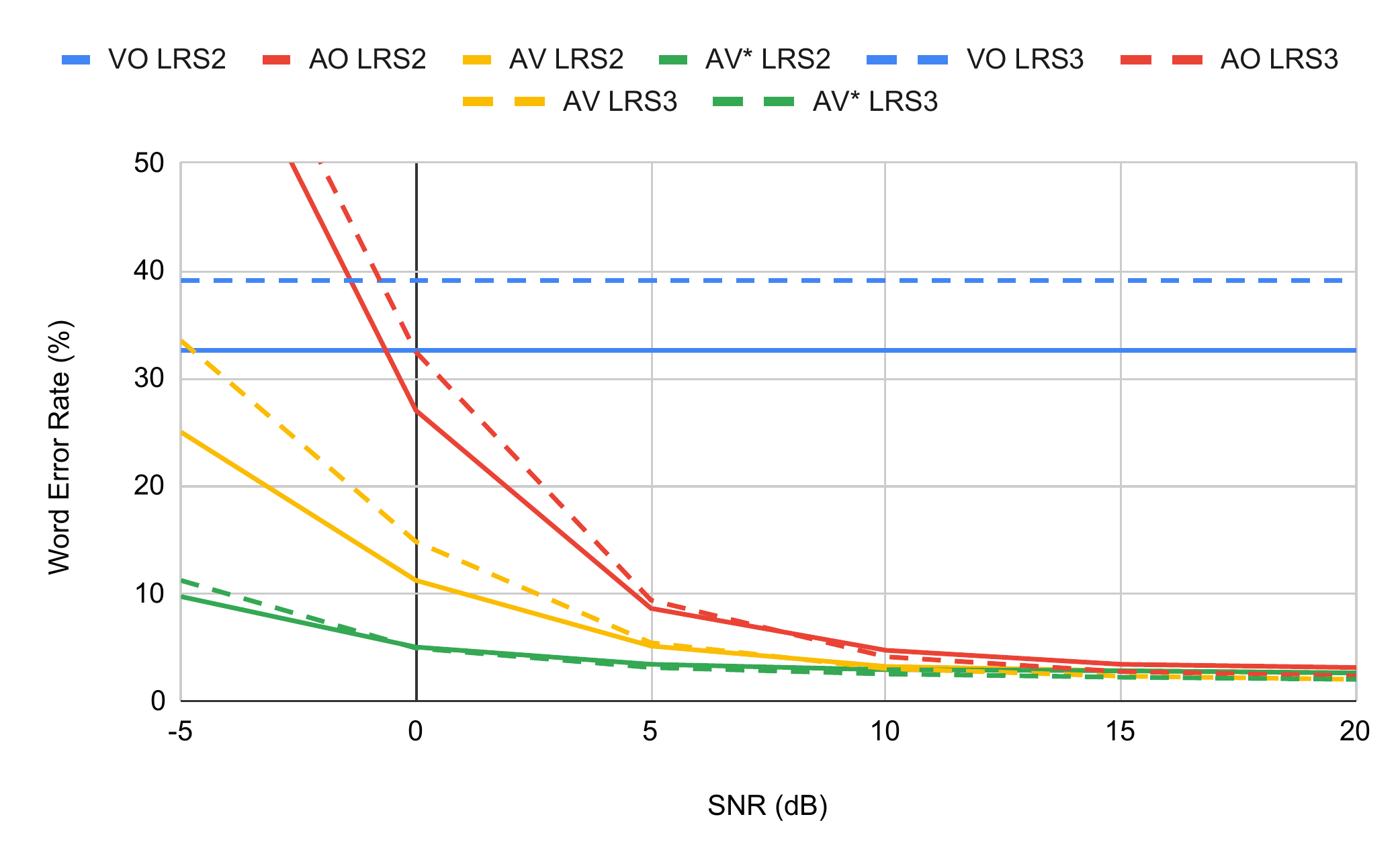}}\\
        \subfloat[White noise]{\includegraphics[width=\textwidth]{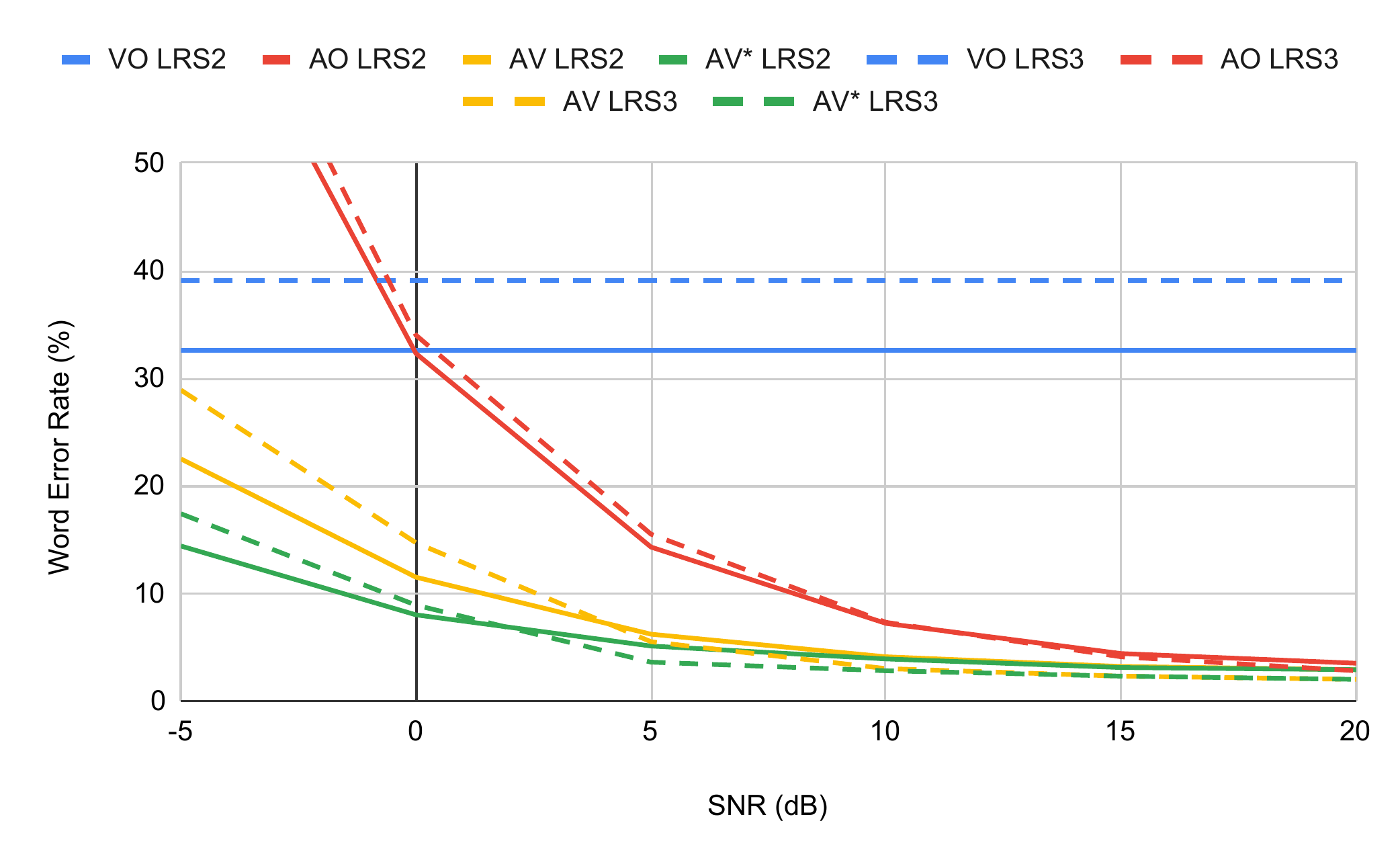}}
        \caption{\textbf{LRS2 and LRS3 test WER (\%) as a function of SNR (dB).} * indicates experiments being trained with babble noise. We measure noise robustness by evaluating our models in the presence of babble and white noise.}
        \label{fig:noise}
    \end{minipage}
\end{figure}

We measure model noise robustness using various types of noise and compare our Audio-Visual Efficient Conformer with recently published methods. Figure~\ref{fig:noise} shows the WER evolution of audio-only (AO), visual-only (VO) and audio-visual (AV) models with respect to multiple Signal to Noise Ratio (SNR) using white noise and babble noise from the NoiseX corpus~\cite{varga1993assessment}. We find that processing both audio and visual modalities can help to significantly improve speech recognition robustness with respect to babble noise. Moreover, we also experiment adding babble noise during training as done in previous works~\cite{petridis2018audio, ma2021end} and find that it can further improve noise robustness at test time. 

\textbf{Robustness to various types of noise.} We gather various types of recorded audio noise including sounds and music. In Table~\ref{table:noise}, we observe that the Audio-Visual Efficient Conformer consistently achieves better performance than its audio-only counterpart in the presence of various noise types. This confirm our hypothesis that the audio-visual model is able to use the visual modality to aid speech recognition when audio noise is present in the input.

\begin{table}[ht]
\centering
\scriptsize
\caption{LRS3 test WER (\%) as a function of SNR (dB).}
\hfill \break
\begin{tabular}{c|ccccccc}
\hline
 \multirow{2}{*}{\textbf{Noise}} & \multirow{2}{*}{\textbf{Mode}} &  \multicolumn{6}{c}{\textbf{SNR (dB)}} \\
& & \textbf{-5} & \textbf{0} & \textbf{5} & \textbf{10} & \textbf{15} & \textbf{20} \\\hline\hline
\multirow{3}{*}{babble} & AO  & 75.9 & 32.4 & 9.3 & 4.1 & 2.7 & 2.3            \\
& AV  & 33.5 & 14.8 & 5.4 & 3.0 & 2.3 & 2.0            \\
& AV*  & 11.2 &	4.9 &	3.1 &	2.5 &	2.2 &	2.0            \\
\hline
\multirow{3}{*}{white} & AO & 77.6 & 34.0 & 15.5 & 7.3 & 4.1 & 2.8           \\
& AV & 28.9 & 14.7 & 5.5 & 3.0 & 2.3 & 2.0          \\
& AV*  & 17.4 &	8.9 &	3.6 &	2.8 &	2.3 &	2.0\\
\hline
\multirow{3}{*}{birds} & AO & 51.8 & 23.9 & 10.9 & 5.9 & 3.7 & 2.8          \\
& AV & 21.6 & 11.5 & 6.2 & 4.1 & 2.9 & 2.4         \\
& AV* & 15.9 &	8.3 &	4.9 &	3.4 &	2.7 &	2.4\\
\hline
\multirow{3}{*}{chainsaw} & AO & 82.9	& 41.2	& 14.8	& 5.5	& 3.7	& 2.7         \\
& AV & 37.8	& 17.3	& 7.6	& 3.9	& 2.6	& 2.3       \\
& AV*  & 25.8 &	10.8 &	5.0 &	3.2 &	2.4 &	2.3 \\
\hline
\multirow{3}{*}{jazz} & AO & 25.3	& 9.7	& 4.1	& 3.1	& 2.6	& 2.3          \\
& AV & 13.9	& 6.0	& 3.2	& 2.4	& 2.3	& 2.0         \\
& AV*  & 10.6 &	4.2 &	2.8 &	2.4 &	2.2 &	2.0\\
\hline
\multirow{3}{*}{\begin{tabular}[c]{@{}c@{}}street \\ raining\end{tabular}} & AO & 58.4	& 23.8	& 8.9	& 4.6	& 3.0	& 2.5         \\
& AV & 27.12	& 10.8	& 5.7	& 3.1	& 2.7	& 2.3        \\
& AV*  & 15.9 &	6.9 &	3.8 &	2.7 &	2.3 &	2.2\\
\hline
\multirow{3}{*}{\begin{tabular}[c]{@{}c@{}}washing \\ dishes\end{tabular}} & AO & 47.8	& 24.5	& 11.5	& 6.0	& 3.7	& 2.8          \\
& AV & 21.3	& 11.5	& 6.1	& 3.6	& 2.8	& 2.3            \\
& AV*  & 14.2 &	7.3 &	4.3 &	2.2 &	2.6 &	2.3\\
\hline
\multirow{3}{*}{train} & AO & 51.3	& 18.6	& 7.0	& 4.0	& 2.9	& 2.5          \\
& AV & 23.1	& 10.1 &	4.7	& 3.0	& 2.4	& 2.2            \\
& AV*  &  14.5 &	6.2 &	3.5 &	2.6 &	2.3 & 2.2\\
\hline
\end{tabular}
\label{table:noise}
\vspace{-1.0cm}
\end{table}

\textbf{Comparison with other methods.} We compare our method with results provided by Ma et al.~\cite{ma2021end} and Petridis~\textit{et al.}~\cite{petridis2018audio} on the LRS2 test set. Table~\ref{table:noise_ma} shows that our audio-visual model achieves lower WER in the presence of babble noise, reaching WER of 9.7\% at -5 dB SNR against 16.3\% for Ma~\textit{et al.}~\cite{ma2021end}.

\begin{table}[ht]
\centering
\scriptsize
\caption{Comparison with Ma~\textit{et al.}~\cite{ma2021end}. LRS2 test WER (\%) as a function of SNR (dB) using babble noise.}
\hfill \break
\begin{tabular}{c|ccccccc}
\hline
 \multirow{2}{*}{\textbf{Method}} & \multirow{2}{*}{\textbf{Mode}} &  \multicolumn{6}{c}{\textbf{SNR (dB)}} \\
& & \textbf{-5} & \textbf{0} & \textbf{5} & \textbf{10} & \textbf{15} & \textbf{20} \\\hline\hline
\multirow{3}{*}{Ma~\textit{et al.}~\cite{ma2021end}} & VO  & 37.9 & 37.9 & 37.9 & 37.9 & 37.9 & 37.9            \\
& AO*  & 28.8& 	9.8 &	7 &	5.2 &	4.5 &	4.2  \\
& AV*  & 16.3 &	7.5 &	6.1 &	4.7 &	4.4 &	4.2            \\
\hline
\multirow{4}{*}{Ours} & VO & 32.6 &	32.6 &	32.6 &	32.6 &	32.6 &	32.6           \\
& AO & 70.5 &	27 &	8.6 &	4.7 &	3.4 &	3.1          \\
& AV  & 25 &	11.2 &	5.1 &	3.2 &	2.8 &	2.6            \\
& AV*  & \textbf{9.7} &	\textbf{5} &	\textbf{3.4} &	\textbf{2.9} &	\textbf{2.8} &	\textbf{2.6}           \\
\hline
\end{tabular}
\label{table:noise_ma}
\vspace{-0.25cm}
\end{table}

\begin{table}[ht]
\centering
\scriptsize
\caption{Comparison with Petridis~\textit{et al.}~\cite{petridis2018audio}. LRS2 test WER (\%) as a function of SNR (dB) using white noise.}
\hfill \break
\begin{tabular}{c|ccccccc}
\hline
 \multirow{2}{*}{\textbf{Method}} & \multirow{2}{*}{\textbf{Mode}} &  \multicolumn{6}{c}{\textbf{SNR (dB)}} \\
& & \textbf{-5} & \textbf{0} & \textbf{5} & \textbf{10} & \textbf{15} & \textbf{20} \\\hline\hline
\multirow{3}{*}{Petridis~\textit{et al.}~\cite{petridis2018audio}} & VO  & 63.5 & 63.5 & 63.5 & 63.5 & 63.5 & 63.5            \\
& AO*  & 85.0 & 45.4 & 19.6 & 11.7 & 9.4 & 8.4  \\
& AV*  & 55.0 & 26.1 & 13.2 & 9.4 & 8.0 & 7.3             \\
\hline
\multirow{4}{*}{Ours} & VO & 32.6 &	32.6 &	32.6 &	32.6 &	32.6 &	32.6           \\
& AO & 73.1 &	32.3 &	14.3 &	7.2 &	4.4 &	3.5          \\
& AV  & 22.5 &	11.5 &	6.2 &	4.1 &	3.2 &	2.9            \\
& AV*  & \textbf{14.4} &	\textbf{8.0} &	\textbf{5.1} &	\textbf{3.9} &	\textbf{3.1} &	\textbf{2.9}           \\
\hline
\end{tabular}
\label{table:noise_petridis}
\vspace{-0.25cm}
\end{table}

\section{Conclusion}

In this paper, we proposed to improve the noise robustness of the recently proposed Efficient Conformer CTC-based architecture by processing both audio and visual modalities. We showed that incorporating multi-scale CTC losses between blocks could help to improve recognition performance, reaching comparable results to most recent autoregressive lip reading methods. We also proposed patch attention, a simpler and more efficient attention mechanism to replace grouped attention in the first audio encoder stage. Our Audio-Visual Efficient Conformer achieves state-of-the-art performance of \reslrstwo\% and \reslrsthree\% on the LRS2 and LRS3 test sets. In the future, we would like to explore other techniques to further improve the noise robustness of our model and close the gap between recent lip reading methods. This includes adding various audio noises during training and using cross-modal distillation with pre-trained
models. We also wish to reduce the visual front-end network complexity without arming recognition performance and experiment with the RNN-Transducer learning objective for streaming applications. 

\section*{Acknowledgments}
This work was partly supported by The Alexander von Humboldt Foundation (AvH).

\clearpage
{\small
\bibliographystyle{ieee_fullname}
\bibliography{egbib}
}

\end{document}